\documentclass{article}
\usepackage[dblblindworkshop, final]{neurips_2025}

\usepackage[utf8]{inputenc} 
\usepackage[T1]{fontenc}    
\usepackage[colorlinks=true, citecolor=blue, urlcolor=blue, linkcolor=blue]{hyperref}
\usepackage{url}            
\usepackage{booktabs}       
\usepackage{amsfonts}       
\usepackage{nicefrac}       
\usepackage{microtype}      
\usepackage{xcolor}         
\usepackage{graphicx}
\usepackage{wrapfig} 
\usepackage{amsmath}

\usepackage{algorithm}
\usepackage{algpseudocode}

\title{A novel approach to classification of ECG arrhythmia types with latent ODEs}
\workshoptitle{Learning from time series for health}

\author{%
  Angelina~Yan\\
  Columbia University\\
  \texttt{ay2672@columbia.edu} \\
\And
  Matt L.~Sampson \\
  Princeton University\\
  \texttt{matt.sampson@princeton.edu} \\
\And 
  Peter~Melchior \\
  Princeton University\\
  \texttt{peter.melchior@princeton.edu} \\
}

\author{%
Angelina~Yan$^{1}$ \quad Matt L.~Sampson$^{2}$ \quad Peter~Melchior$^2$  \\
$^1$Columbia University \quad $^2$ Princeton University\\
\texttt{ay2672@columbia.edu}\\
\texttt{\{matt.sampson, peter.melchior\}@princeton.edu}
}

\begin{document}
\algrenewcommand\algorithmicrequire{\textbf{Input:}}   
\algrenewcommand\algorithmicensure{\textbf{Return:}}   

\maketitle


\begin{abstract}
12-lead ECGs with high sampling frequency are the clinical gold standard for arrhythmia detection, but their short-term, spot-check nature often misses intermittent events. Wearable ECGs enable long-term monitoring but suffer from irregular, lower sampling frequencies due to battery constraints, making morphology analysis challenging. We present an end-to-end classification pipeline to address these issues. We train a latent ODE to model continuous ECG waveforms and create robust feature vectors from high-frequency single-channel signals. We construct three latent vectors per waveform via downsampling the initial 360 Hz ECG to 90 Hz and 45 Hz. We then use a gradient boosted tree to classify these vectors and test robustness across frequencies. Performance shows minimal degradation, with macro-averaged AUC-ROC values of 0.984, 0.978, and 0.976 at 360 Hz, 90 Hz, and 45 Hz, respectively, suggesting a way to sidestep the trade-off between signal fidelity and battery life. This enables smaller wearables, promoting long-term monitoring of cardiac health.
\end{abstract}

\section{Introduction}
Cardiovascular diseases (CVDs) are one of the leading causes of death worldwide \citep{martin20252025}. Many CVDs manifest in arrhythmias, which are routinely captured by electrocardiograms (ECGs), the standard tool in clinical settings. The morphological characteristics of the ECG waveform reflect normal versus abnormal heart beats that can be classified into distinct categories. Classifying heart beats is the basis for the diagnosis of arrhythmias and underlying heart problems \citep{acharya2017automated}.

Traditionally, 12-lead ECGs are considered the gold standard for diagnosis and offer the most spatially comprehensive view of the electrical activities of the heart \citep{schlant1992guidelines}. However, the 12-lead ECG acquisition process requires trained professionals and provides only a short-term, spot-check measurement. In contrast, modern wearable ECGs (typically single-channel) require no additional setup and do not disrupt daily life \citep{bouzid2022remote}. The shift in public health goals from reactive treatment to proactive prevention has led to a rapid increase in the popularity of wearables. Their ease of use has made them ideal for long-term, continuous monitoring and early detection of paroxysmal or asymptomatic events \citep{babu2024wearable}. For instance, wearable patches can be worn for weeks at a time and automatically acquire signals, producing continuous waveforms that can detect abnormalities and deviations from individual baselines not present in short-term 12-lead ECG waveforms \citep{sana2020wearable}. This capability is crucial for closing the diagnostic gap in detecting infrequent or transient cardiac arrhythmias.

Despite their advantages, wearable ECGs often suffer from low sampling frequencies in the interest of battery conservation. This makes accurate morphology analysis from wearable data challenging since low sampling frequencies compromise signal fidelity \citep{kwon2018electrocardiogram}. In this work, we present an end-to-end classification pipeline robust to sampling frequency variations. This is achieved by training a latent ODE on high-frequency 360 Hz ECG signals, which handles both potential signal irregularity and noise. It models the underlying continuous signal as a differential equation in latent space \citep{rubanova2019latent}. As part of this generative modeling process, the latent ODE trains an encoder that maps each continuous time-series into a single latent vector. This encoding process has been shown to be effective for creating automated feature vectors to then be used for classification in scientific domains \citep{sampson2025path}.
To perform arrhythmia classification, the latent feature vectors are then fed into a gradient boosted decision tree \citep{friedman2001greedy}. Importantly, the trained encoder model of the latent ODE is able to create informative latent feature vectors from both low and high frequency data.

\subsection*{Background and related work}

Given the immense volume of waveform data produced by wearables, it is infeasible to solely rely on the availability of cardiologists for diagnosis. In addition, manually interpreting subtle visual changes in ECG waveforms is challenging, dependent on physician experience, and subject to interpretative discrepancies \citep{ansari2023deep}. Fortunately, modern computational predictors have shown promise in assisting clinical decision-making. Over the last decade, the field of computational ECG analysis has undergone a paradigm shift from manual feature engineering to end-to-end deep learning models with fully automated feature extraction \citep{montenegro2022human}. Recent work has demonstrated that modern deep learning neural network (DNN) models can not only assist but even rival or outperform human experts in the detection and classification of arrhythmias from 12-lead ECGs and single-lead ambulatory ECGs \citep{ribeiro2020automatic, hannun2019cardiologist}. We present an alternative approach in end-to-end modeling, where we automate the creation of the feature vectors yet allow for a wide variety of downstream classification algorithms to be used.

\section{Methodology}
Here we describe the training dataset as well as our multi-step classification routine. We report further details on model architecture and training routines for all models in \autoref{sec:appendix}.

\subsection{Training data}
We obtained the ECG heart beat signals from the widely used open access PhysioNet MIT-BIH Arrhythmia Database \citep{932724, goldberger2000physiobank}. The database consists of 48 two-channel ECG recordings taken from 47 individuals at a sampling rate of 360 Hz. For this work, we chose the modified limb lead II (MLII) data for each recording because many wearables also acquire single-lead ECG data \citep{bouzid2022remote}. We performed a denoising and peak matching routine on the raw ECG beats as described in \citet{liu2020ecg}. We follow the Advancement of Medical Instrumentation (AAMI) standard for the classification of each beat into one of five classes: normal (N), ventricular (V), supraventricular (S), fusion of normal and ventricular (F), and unknown beats (Q). A total of 88887 ECG beats were extracted from this dataset. For the classification testing, we perform a naive downsampling of the ECG data by taking every $n$th entry to achieve the desired downsampled frequency. The downsampled data is then used for classification by our latent-ODE/decision tree pipeline shown in \autoref{alg:full_pipeline}.

\subsection{Path-minimized latent ODEs}
We train a latent ODE \citep{rubanova2019latent} on the ECG time-series data in the MIT-BIH dataset. Specifically, we use a path-minimized latent ODE as described in \citep{sampson2025path}, which replaces the common variational component from the loss with an $\ell_2$ regularizer that acts to minimize the point-to-point distance within a single latent trajectory. This has been shown to improve generative fidelity and inference performance of classifiers trained on the latent encodings. The latent ODE modeling has two goals: to provide a generative model of the full ECG dataset and to perform a robust encoding of each time-series into a low-dimensional latent vector $\mathbf{z}_0$, which we then use for classification.

\subsection{Gradient boosted decision tree} To perform the arrhythmia classification, we use a gradient boosted decision tree (GBDT) \citep{friedman2001greedy}. This classifier is trained on the encoded latent vectors, $\mathbf{z}_0$, of the original ECG segments and their class labels. A benefit of this multi-step approach is that by training the GBDT on the latent vectors instead of features directly from the ECG itself, we leverage the latent ODE model to produce a latent vector that closely matches a higher quality signal, even if the original ECG signal were of lower quality due to noise, sampling frequency, or other forms of corruption. 

We perform a synthetic minority over-sampling technique (SMOTE)  \citep{chawla2002smote} on the latent feature vectors such that we have the same number of samples in each of the 5 classes during the \textit{training} of the GBDT classifier. Our latent feature vectors are conditionally sampled from the latent ODE encoder. During \textit{testing}, we take $n$ random latent vectors for each ECG and select the mode of the $n$ random samples
to be the final classification. 

\subsection{Full classification pipeline}
We present pseudocode for the full classification routine in \autoref{alg:full_pipeline}.

\begin{algorithm}[H]
\caption{Arrhythmia Classification Pipeline}
\label{alg:full_pipeline}
\begin{algorithmic}[1]
\Require ECG time-series vector $(x,t)$; trained latent ODE; trained gradient boosted decision tree classifier (GBDT); ensemble size $n$, a random seed sampling  
\Ensure Final class prediction $y^\star$

\Function{Predict}{$x, t, n$}
    \For{$i = 1$ to $n$}
    \State $\mathbf{z_{0,i}} \gets \Call{LatentODE}{x,t, \rm{seed}_i}$ \Comment{Encoding step}
    \State ${\hat{y}}.\rm{append}\left(\Call{GBDT}{\mathbf{z_{0,i}}}\right)$ \Comment{GBDT predicts one label per $z_{0,i}$}
    \EndFor
    \State $y^\star \gets \operatorname{mode}(\hat{y})$ \Comment{Final class is the majority vote over $\{\hat{y}_i\}_{i=1}^n$}
    \State \textbf{return} $y^\star$
\EndFunction

\end{algorithmic}
\end{algorithm}

\section{Experiments and Results}
We perform a set of arrhythmia classification experiments on the MIT-BIH dataset using our latent ODE to GBDT classifier routine. We split our dataset into training/validation/test sets with a 70/15/15 split, respectively. All results reported come from the model with the best validation performance re-evaluated on the test set.

\subsection{Latent ODE performance}
\begin{figure}[h]
    \centering
    \vspace{0em} 
    \includegraphics[width=0.99\textwidth]{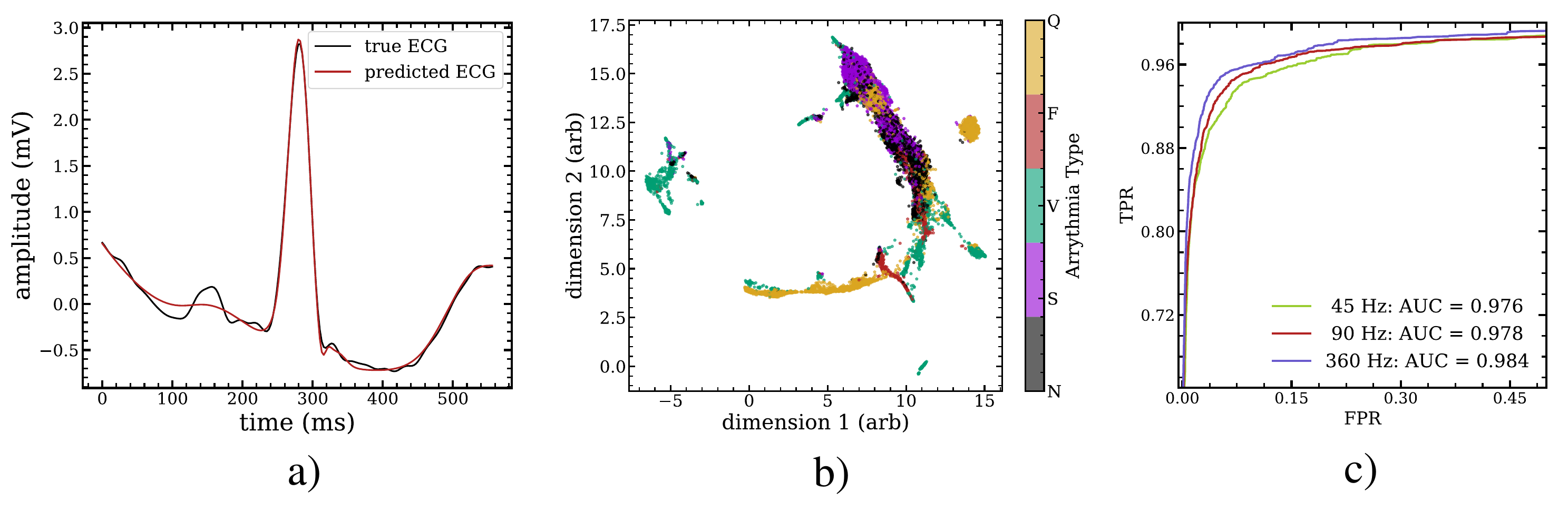}
    \caption{\textbf{a)} Reconstruction of a randomly sampled ECG timeseries with the latent ODE prediction in red and the true signal in black. \textbf{b)} UMAP of the latent feature vectors from the test set of the BIH-MIT data, with colors indicating the arrhythmia class. \textbf{c)} AUC-ROC curves from the GBDT classifier based on latent ODE encodings of ECG curves sampled at 360, 90, and  45 Hz.}
    \label{fig:ecg}
    \vspace{-0.5em}
\end{figure}

We present a sample result from the trained latent ODE in panel \textbf{a)} of \autoref{fig:ecg}. By using only 45 latent dimensions, we see excellent agreement between the true (black) and sampled (red) ECG curve, indicating the model's  ability to learn the underlying structure of the ECG signals, while ignoring potential minor signal corruptions. We present additional samples in \autoref{sec:additional}.

In panel \textbf{b)} we show a UMAP \citep{mcinnes2018umap} of the test set of encoded ECG timeseries vectors, where each point is colored by the arrhythmia class. We see a clear structure in the projection with similar class labels clustering together in this reduced latent space. This give confidence that the latent ODE is providing meaningful encodings for the different types of ECG signals observed.  

\subsection{Classification results}
We show the classification results in \autoref{tab:results}, displaying accuracy, precision, recall, and F1 score for all 5 classes. We report the classification results from the ECG series at sampling frequencies of 360 Hz, 90 Hz, and 45 Hz. We also show a set of one-vs-all AUC-ROC plots in panel \textbf{c)} of \autoref{fig:ecg} for all three frequencies. For the multiclass classification setting, predicted class labels were determined by selecting the class with the highest predicted probability from the gradient boosted decision tree. Performance metrics (accuracy, precision, recall, and F1) were computed from these predicted labels using \texttt{scikit-learn}’s default implementations with the prediction threshold set to 0.5.

\begin{table}[h]
\centering
\small
\caption{Classification results for ECG data sampled at 360 Hz, 90 Hz, and 45 Hz. We show the per-class accuracy, precision, recall and F1 scores. We also report the macro-averaged results for each sampling frequency in the bottom row.}
\vspace{-0.5em}
\resizebox{\textwidth}{!}{
\begin{tabular}{c c | ccc | ccc | ccc | ccc}
\toprule
 & & \multicolumn{3}{c|}{Accuracy ($\%$)}  & \multicolumn{3}{c|}{Precision} & \multicolumn{3}{c|}{Recall} & \multicolumn{3}{c}{F1} \\ 
\cmidrule(lr){3-5} \cmidrule(lr){6-8} \cmidrule(lr){9-11} \cmidrule(lr){12-14}
class & count & $360\ \rm{ Hz}$  & $90\ \rm{ Hz}$ & $45\ \rm{ Hz}$ & $360\ \rm{ Hz}$ & $90\ \rm{ Hz}$ & $45\ \rm{ Hz}$ & $360\ \rm{ Hz}$ & $90\ \rm{ Hz}$ & $45\ \rm{ Hz}$ & $360\ \rm{ Hz}$ & $90\ \rm{ Hz}$ & $45\ \rm{ Hz}$ \\ 
\midrule   
N & 10988 & 98.0 & 97.7 & 97.7 & 0.98 & 0.98 & 0.98 & 0.98 & 0.98 & 0.98 & 0.98 & 0.98 & 0.98 \\
S & 327   & 75.2 & 70.5 & 70.1 & 0.69 & 0.66 & 0.64 & 0.75 & 0.70 & 0.71 & 0.72 & 0.68 & 0.67 \\
V & 918   & 93.9 & 93.6 & 92.6 & 0.93 & 0.90 & 0.91 & 0.94 & 0.94 & 0.93 & 0.93 & 0.92 & 0.92 \\
F & 99    & 72.3 & 69.6 & 60.8 & 0.72 & 0.75 & 0.65 & 0.72 & 0.70 & 0.61 & 0.72 & 0.72 & 0.63 \\
Q & 1001  & 95.4 & 94.3 & 93.3 & 0.95 & 0.93 & 0.92 & 0.95 & 0.94 & 0.93 & 0.95 & 0.94 & 0.93 \\ 
macro-avg & -- & 87.0 & 85.9 & 82.9 & 0.85 & 0.84 & 0.82 & 0.87 & 0.85 & 0.83 & 0.86 & 0.85 & 0.82\\
\bottomrule
\end{tabular}
}
\label{tab:results}
\vspace{-1.0em}
\end{table}

\begin{figure}[h]
    \centering
    \includegraphics[width=0.95\linewidth]{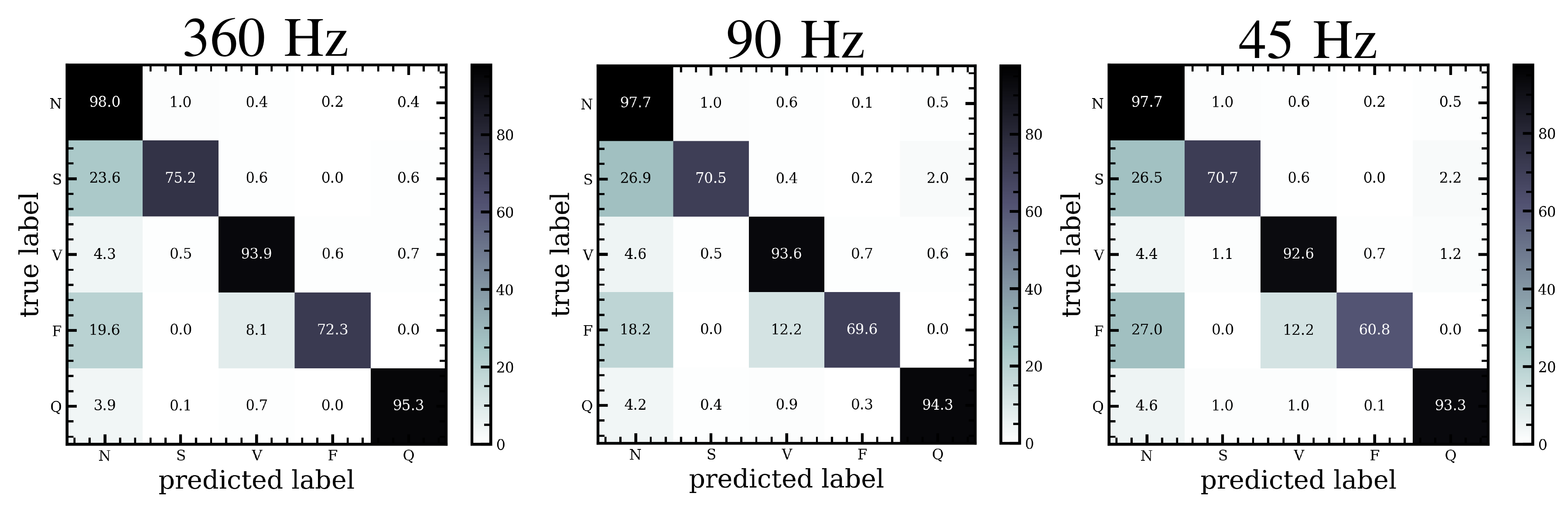}
    \caption{Normalized confusion matrices from the GBDT on the ECG data at samping frequencies of 360 Hz, 90 Hz, and 45 Hz from left to right, respectively.}
    \label{fig:confusion}
\end{figure}

In \autoref{fig:confusion} we show normalized confusion matrices for the GBDT performance on the test data for the 360 Hz, 90 Hz, and 45 Hz trials, respectively. From this we see the minority classes S and F are most commonly mislabelled as N, with this issue becoming particularly clear in the 45 Hz trial.

\section{Discussion}
The primary advantage of our classification pipeline is its inherent robustness to the challenges posed by real-world ECG data, specifically sparsity, incompleteness, and low sampling frequency. We empirically demonstrate this robustness in \autoref{tab:results}, which details our model's per-class classification metrics (accuracy, precision, recall, and F1 scores) across data sampled at 360 Hz, 90 Hz, and 45 Hz as well as the macro-averaged results for each metric. The results show a minimal degradation in performance as the sampling rate decreases, indicating our pipeline's ability to construct high-quality feature vectors even with a significantly reduced data stream. 

We note that the difference (for all sampling frequencies) between the relatively high AUC-ROC scores of 0.984, 0.978, and 0.976 and the lower macro-averaged F1 scores in \autoref{tab:results} of 0.86, 0.85, and 0.82 hints that our model still suffers from class imbalance issues, with the relatively poor performance on the minority classes reducing the macro-averaged F1 scores.
While our pipeline shows promise as a novel method for treating irregular and low frequency ECG data, its current performance is limited by the relatively small size of the MIT-BIH dataset. In future work, we hope to improve our results by training on larger datasets with a greater representation of the undersampled S and F classes. The main quantitative difference we note here is: the accuracy for the S class decreases from $75.2\%$ at $360\ \rm{Hz}$ to $70.1\%$ at $45\ \rm{Hz}$, and the accuracy for the F class decreases from $72.3\%$  at $360\ \rm{Hz}$ to $60.8\%$ at $45\ \rm{Hz}$. Larger datasets as well as alternate oversampling techniques may help alleviate these issues in future efforts.

\section{Limitations and future work}
We point out this study represents preliminary work in this area. A clear limitation is the lack of practical testing of this new classification pipeline on edge devices, which would allow for accurate estimates of inference and memory costs. Another limitation is the relatively limited dataset used. We perform all testing on the publicly available MIT-BIH dataset, which contains significant class imbalance, a total of only 47 individuals, and is not a wearable dataset. In future work, we plan on exploring the performance of this pipeline over a larger wearable dataset containing raw ECG signals.

\section{Conclusion}
By combining a latent ODE model with a decision tree classifier, we demonstrate that high-quality classification is achievable at lower frequencies. Our work enables the use of smaller, less power-hungry sensors with longer battery life and the development of smaller, more comfortable wearable devices. Extending battery life also increases the length of continuous monitoring that can be provided by a wearable, which catches more infrequent or transient pathologies. Taken together, these improvements could encourage more patients to adopt long-term monitoring, aligning with the modern attitude shift toward a proactive approach to public health.

\bibliography{refs}
\bibliographystyle{plainnat}

\appendix

\section{Model details}
\label{sec:appendix}

\subsection{Latent ODE details}
Our latent ODE-RNN architecture is implemented in \texttt{jax} and \texttt{equinox} \citep{jax2018github, kidger2021equinox}. We use a feed forward neural network to model our ODE function as in \citep{rubanova2019latent}. We use Tanh activation functions with 2 layers and a width of 50.
We use ODE solvers from the \texttt{diffrax} package \citep{kidger2021on}, specifically the 5th order Tsit5() solver, Tsitouras 5/4 method (5th order Runge-Kutta), with adaptive steps and an initial dt = 0.001. We use 45 hidden and 45 latent dimensions and train for 50000 steps. We follow the implementation from \citet{sampson2025path} for the path-regularization. The training time is approximately 2 hours on a single NVIDIA A100.

\subsection{Boosted classifier details}
We use a gradient boosted classifier from the \texttt{scikit-learn} package \citep{pedregosa2011scikit}. We train with 1000 trees and a max depth of 8. We find increasing the depth tends to bias the classifier towards the majority class during validation even with equal sized training classes during training (from SMOTE oversampling).

\section{Additional results}
\label{sec:additional}
We show additional results here for both the classification performance and the latent ODE modeling.

\begin{figure}[h]
    \centering
    \includegraphics[width=0.95\linewidth]{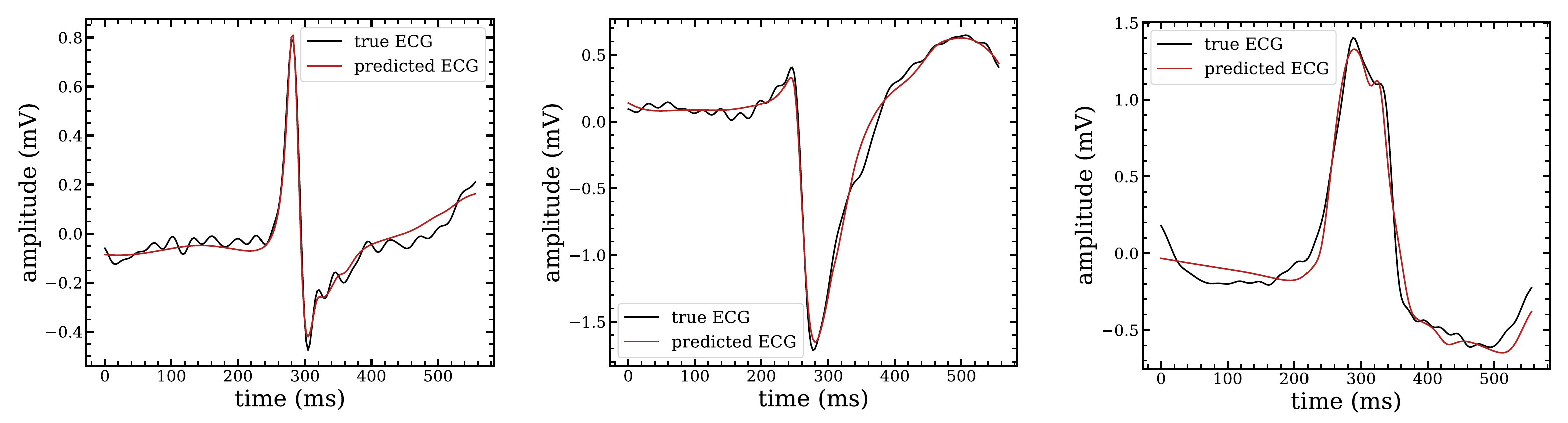}
    \caption{Additional reconstructions of randomly sampled ECG timeseries with the latent ODE prediction in red and the true signal in black.}
    \label{fig:ecgs}
\end{figure}

We show some extra random samples of reconstructed ECG time-series from the trained latent ODE in \autoref{fig:ecgs}. We can see that the trained model accurately reconstructs a wide variety of ECG shapes.

\end{document}